\documentclass[sigconf]{acmart}

\AtBeginDocument{%
  }

\copyrightyear{2023}
\acmYear{2023}
\setcopyright{acmlicensed}

\acmConference[MM '23] {Proceedings of the 31st ACM International Conference on Multimedia}{October 29--November 3, 2023}{Ottawa, ON, Canada.}
\acmBooktitle{Proceedings of the 31st ACM International Conference on Multimedia (MM '23), October 29--November 3, 2023, Ottawa, ON, Canada}
\acmBooktitle{Proceedings of the 31st ACM International Conference on Multimedia (MM '23), October 29--November 3, 2023, Ottawa, ON, Canada}
\acmPrice{15.00}
\acmISBN{979-8-4007-0108-5/23/10}
\acmDOI{10.1145/3581783.3612487}

\usepackage{amsmath}
\usepackage{enumitem}
\usepackage{subcaption}
\usepackage{multirow}
\usepackage{pifont}

\newcommand{\cmark}{\ding{51}}%
\newcommand{\xmark}{\ding{55}}%
\newcommand{\myparagraph}[1]{\vspace{0.1\baselineskip}\noindent{\textbf{#1.}}~}

\begin{document}

\title[Online Multi-view Anomaly Detection with Disentangled Product-of-Experts Modeling]{Debunking Free Fusion Myth: Online Multi-view Anomaly Detection with Disentangled Product-of-Experts Modeling}


\author{Hao Wang}
\email{cshaowang@gmail.com}
\affiliation{%
  \institution{Southwest Jiaotong University}
  \country{}
}

\author{Zhi-Qi Cheng}
\email{zhiqic@cs.cmu.edu}
\affiliation{
  \institution{Carnegie Mellon University}
  \country{}
}

\author{Jingdong Sun}
\email{jingdons@andrew.cmu.edu}
\affiliation{
  \institution{Carnegie Mellon University}
  \country{}
}

\author{Xin Yang}
\authornote{Corresponding author.}
\email{yangxin@swufe.edu.cn}
\affiliation{%
  \institution{Southwestern University of Finance and Economics}
  \country{}
}

\author{Xiao Wu}
\email{wuxiaohk@swjtu.edu.cn}
\affiliation{%
  \institution{Southwest Jiaotong University}
  \country{}
}

\author{Hongyang Chen}
\email{hongyang@zhejianglab.com}
\affiliation{%
  \institution{Zhejiang Lab}
  \country{}
}

\author{Yan Yang}
\authornotemark[1]
\email{yyang@swjtu.edu.cn}
\affiliation{%
  \institution{Southwest Jiaotong University}
  \country{}
}

\renewcommand{\shortauthors}{Hao Wang et al.}

\begin{abstract}
Multi-view or even multi-modal data is appealing yet challenging for real-world applications. Detecting anomalies in multi-view data is a prominent recent research topic. However, most of the existing methods 1) are only suitable for two views or type-specific anomalies, 2) suffer from the issue of fusion disentanglement, and 3) do not support online detection after model deployment. To address these challenges, our main ideas in this paper are three-fold: multi-view learning, disentangled representation learning, and generative model. To this end, we propose dPoE, a novel multi-view variational autoencoder model that involves (1) a Product-of-Experts (PoE) layer in tackling multi-view data, (2) a Total Correction (TC) discriminator in disentangling view-common and view-specific representations, and (3) a joint loss function in wrapping up all components. In addition, we devise theoretical information bounds to control both view-common and view-specific representations. Extensive experiments on six real-world datasets demonstrate that the proposed dPoE outperforms baselines markedly.
\end{abstract}



\maketitle

\section{Introduction}

\begin{table*}[!t]
    \small
    \centering
    \caption{\small A summary of existing methods and our method. The symbol \cmark~ denotes such a method is for this issue. Opposite to this, \xmark~ denotes \textit{not for}. The symbol $*$ denotes that the method needs to store the hidden representation of the training instances as the defined anomaly score depends on k-nearest neighbors. As noted in this table, our model can tackle all issues.}
    \begin{tabular}{cc|ccccccc}
    \toprule
        \multirow{2}{*}{\textsc{Learning Mode}} & \multirow{2}{*}{\textsc{Method}} & \multicolumn{4}{c}{\textsc{Issue-1}} & \multirow{2}{*}{\textsc{Issue-2}} & \multirow{2}{*}{\textsc{Issue-3}} \\
        & & \textsc{Type-I} & \textsc{Type-II} & \textsc{Type-III} & >\textsc{Two Views} & & \\
    \addlinespace[2pt]
    \midrule
    \addlinespace[2pt]
         \multirow{9}{*}{\textsc{Non-deep}} & HOAD (\citet{DBLP:conf/icdm/GaoFTPH11}) & \xmark & \cmark & \xmark & \xmark & \xmark & \xmark \\
         & AP (\citet{DBLP:conf/cikm/AlvarezYKI13}) & \xmark & \cmark & \xmark & \xmark & \xmark & \xmark \\ 
         & DMOD (\citet{DBLP:conf/ijcai/ZhaoF15a}) & \cmark & \cmark & \xmark & \xmark & \xmark & \xmark \\ 
         & MLRA (\citet{DBLP:conf/sdm/LiSF15}) & \cmark & \cmark & \xmark & \xmark & \xmark & \xmark \\ 
         & PLVM (\citet{DBLP:conf/nips/IwataY16}) & \xmark & \cmark & \xmark & \cmark & \xmark & \cmark \\
         & LDSR (\citet{DBLP:conf/aaai/Li0DZF18}) & \cmark & \cmark & \cmark & \cmark & \xmark & \xmark \\ 
         & CL (\citet{DBLP:conf/aaai/GuoZ18}) & \xmark & \cmark & \xmark & \xmark & \xmark & \xmark \\ 
         & MuvAD (\citet{DBLP:conf/aaai/ShengZL019}) & \cmark & \cmark & \xmark & \xmark & \xmark & \xmark \\ 
         & HBM (\citet{DBLP:conf/ijcai/WangL20}) & \cmark & \cmark & \cmark & \cmark & \xmark & \cmark \\ 
         & SRLSP (\citet{DBLP:journals/tkdd/WangCLFZZ23}) & \cmark & \cmark & \cmark & \cmark & \xmark & ~\cmark* \\ 
         \addlinespace[2pt]
         \midrule
         \addlinespace[2pt]
         \multirow{3}{*}{\textsc{Deep}} & MODDIS (\citet{DBLP:conf/icdm/JiHHWXSL19}) & \cmark & \cmark & \cmark & \cmark & \xmark & ~\cmark* \\
         & NCMOD (\citet{DBLP:conf/aaai/Cheng0L21}) & \cmark & \cmark & \cmark & \cmark & \xmark & ~\cmark* \\
         & dPoE (Our model) & \cmark & \cmark & \cmark & \cmark & \cmark & \cmark \\
    \addlinespace[2pt]
    \bottomrule
    \end{tabular}
    \label{related_work_summary}
\end{table*}

Anomaly detection  (a.k.a outlier detection) is a critical and challenging task in many real-world applications, such as fraud detection~\cite{DBLP:conf/cikm/DouL0DPY20}, network intrusion identification~\cite{DBLP:journals/tkde/LiYZ19}, and video event surveillance~\cite{DBLP:conf/mm/YuWCZXYK20}. Typically, anomaly detection assumes that an instance is considered anomalous if its pattern significantly deviates from the majority. These approaches are commonly implemented using clustering-based methods~\cite{DBLP:conf/nips/ChenA018}, neighborhood-based methods~\cite{DBLP:conf/nips/GuAR19}, or reconstruction-based methods~\cite{DBLP:conf/iclr/ZongSMCLCC18}. However, with the prevalence of multi-view data in today's world, where instances are described by multiple views or modalities, such as different news organizations reporting the same news or an image being encoded by various features, traditional anomaly detection methods that focus on single-view data are insufficient. Multi-view data provide complementary information for an instance, and all views share consensus information for that instance, which make (i) the data distribution in each view complex and (ii) the characteristics across different views inconsistent.

\begin{figure}[!htbp]
    \centering
    \includegraphics[width=\linewidth]{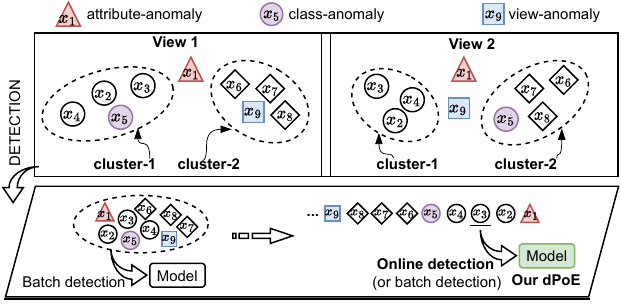}
    \caption{\small Illustration of three types of anomalies (\textit{upper}), and batch detection \textit{versus} online detection (\textit{bottom}). Most existing models are for one or two types of anomalies in the batch detection manner. Our model is for all three types of anomalies in the batch or online detection manner.}
    \label{anomaly_type}
\end{figure}

As introduced in \cite{DBLP:conf/aaai/Li0DZF18}, multi-view data could have three types of anomalies.
Figure \ref{anomaly_type} illustrates these three types of anomalies in multi-view data.
Specifically, the scenario involves nine instances $x_i~(i=1,...,9)$ with two views, assuming two patterns (e.g., clusters) in each view.
This scenario includes an instance $x_1$ that does not belong to any clusters in both views, which is referred to as an \textit{attribute-anomaly}.
Also in this scenario, instance $x_5$ belongs to cluster-1 in view 1 but to cluster-2 in view 2, which is referred to as a \textit{class-anomaly}.
Finally, instance $x_9$ should not belong to cluster-2 in view 1 and does not belong to any clusters in view 2, referred to as a \textit{view-anomaly}.
Following this case, we can categorize multi-view anomalies into three types:
\begin{enumerate}[label=\textsc{Type-\Roman*}:,align=left]
    \item \textit{Attribute-anomaly is an extension of the conventional anomalies in single-view data. It refers to that the pattern of an instance deviates significantly from the others in every view.}
    \item \textit{Class-anomaly refers to that the pattern of an instance in some or even all views is inconsistent.}
    \item \textit{View-anomaly refers to that the pattern of an instance abnormally follows the others in certain views and deviates significantly from the others in other views.}
\end{enumerate}

Anomaly detection on multi-view data (or simply \textit{multi-view anomaly detection}) receives certain attention in recent years. HOAD \cite{DBLP:conf/icdm/GaoFTPH11} is an early multi-view anomaly detection method for \textsc{Type-II} anomaly detection across two views. DMOD \cite{DBLP:conf/ijcai/ZhaoF15a} aims to identify both \textsc{Type-I} and \textsc{Type-II} anomalies in multi-view data. LDSR \cite{DBLP:conf/aaai/Li0DZF18} raises the problem of detecting three types of anomalies. So now the goal of multi-view anomaly detection is not only to identify
instances that deviate from normal patterns (i.e., \textsc{Type-I} and \textsc{Type-III} anomalies), but also to identify
instances with inconsistent behavior among multiple views (i.e., \textsc{Type-II} anomalies). What's more, there are three major issues (\textsc{\textbf{Issue-1}}: \textit{model versatility}, \textsc{\textbf{Issue-2}}: \textit{fusion disentanglement}, and \textsc{\textbf{Issue-3}}: \textit{online detection}) to be resolved for multi-view anomaly detection. Table~\ref{related_work_summary} is a summary of the existing methods and our method. Based on this table and Figure~\ref{anomaly_type}, we expatiate each of the three issues as follows: 
\begin{enumerate}[label=\textsc{Issue-\arabic*}:,itemindent=26pt,leftmargin=*]
    \item Detecting three types of anomalies holistically in a single model is nontrivial for multi-view (more than two views) data. We refer to this issue as \textit{model versatility}. 
    However, most existing methods (e.g., HOAD, AP, DMOD, MLRA, PLVM, CL, and MuvAD) only concern one or two types of anomalies,
    and most (e.g., HOAD, AP, MLRA, CL, and MuvAD) are designed for two-view data. In addition, most of them are non-deep learning methods.
    \item Disentanglement of fusion (view-common representation) and individuation (view-specific representation) is essential for multi-view anomaly detection. We refer to this issue as \textit{fusion disentanglement}. 
    However, existing methods fuse all views directly (e.g., PLVM, LDSR, MuvAD, HBM, SRLSP, MODDIS, and NCMOD) or treat each view separately (e.g., HOAD, AP, DMOD, MLRA, and CL).
    \item Online detection is crucial for an anomaly detection model after the model is deployed into a real-world application. 
    Existing work mostly assumes the batch detection setting,
    which is too strong in practice. A more practical setting is the online scenario where a new (possibly never-ending) instance is submitted to the model for detection after model deployment. Although some existing methods (e.g., SRLSP, MODDIS, and NVMOD) are suited to online detection, they need to store the hidden representation of previous instances because their anomaly score function depends on k-nearest neighbors. PLVM and HBM are also suited to online detection, but they are not for \textsc{Issue-2}.
\end{enumerate}

\myparagraph{Motivation}
As expatiated above, there is no existing work that can tackle all three issues holistically using a single model. The existing work only concerns one or two of the above issues. However, it is excessive and time-consuming to build a standalone model for each of these issues. In this work, we aim to deal with all these issues using a holistic model. To this end, we propose a \underline{d}isentangled \underline{P}roduct-\underline{o}f-\underline{E}xperts (dPoE) model.

\myparagraph{Main Ideas}
Our main ideas here are three-fold: \textit{multi-view learning} (for \textsc{Issue-1}), \textit{disentangled representation learning} (for \textsc{Issue-2}), and \textit{generative model} (for \textsc{Issue-3}). Specifically, 
to address \textsc{Issue-1}, we propose a new multi-view variational autoencoder framework with tight bounds (\textbf{Section~\ref{sec3.1}}) and further saddle this framework with a novel Product-of-Experts (PoE) layer (\textbf{Section~\ref{sec3.2}}). PoE is an ensemble technique that integrates multiple probability distributions (``experts'') by multiplying their probabilities together \cite{DBLP:journals/neco/Hinton02}. In our case, we assign each view an expert and then combine the detection results from all experts as the final decision on a test instance.
To address \textsc{Issue-2}, we propose to disentangle view-common representation from view-specific representations using a Total Correction (TC) discriminator (\textbf{Section~\ref{sec3.3}}). TC is an information measure of dependence for a group of
random variables \cite{DBLP:journals/ibmrd/Watanabe60,DBLP:conf/icml/KimM18}. In this work, we design a discriminator to minimize the TC value of view-common and view-specific representations. To address \textsc{Issue-3}, we formulate a new loss function to wrap up all components of dPoE in a generative model manner such that our model supports online anomaly detection via data inference (\textbf{Section~\ref{sec3.4}}).

In summary, we make the following contributions:
\begin{itemize}[leftmargin=*]
    \item We propose a novel generative model (called dPoE) for multi-view anomaly detection that considers model versatility, fusion disentanglement, and online detection. To the best of our knowledge, this work is the first to tackle three types of anomalies using a holistic model for online multi-view anomaly detection.
    \item We propose a new multi-view VAE framework with tight bounds capped on both view-common and view-specific representations. We saddle the framework with a Product-of-Experts (PoE) layer and a Total Correlation (TC) discriminator for disentangled representation learning. We formulate a joint loss for dPoE. 
    \item We evaluate our dPoE on six real-world datasets. Extensive experiments demonstrate that the proposed dPoE outperforms the state-of-the-art counterparts.
\end{itemize}

\section{Preliminaries}\label{beta_vae}
In this section, we introduce the motivation of disentangled representation learning together with the $\beta$-VAE~\cite{DBLP:conf/iclr/HigginsMPBGBML17}, which is a basic component of our model.

Disentangled representation learning aims to identify and disentangle the underlying independent factors hidden in the observed data for various downstream tasks \cite{DBLP:journals/pami/BengioCV13}. $\beta$-VAE \cite{DBLP:conf/iclr/HigginsMPBGBML17} is a typical model for disentangled representation learning.
It is a modification of the vanilla VAE \cite{DBLP:journals/corr/KingmaW13} with a special emphasis to discover disentangled latent factors in an unsupervised manner. $\beta$-VAE follows the same architecture of VAE and has an adjustable hyperparameter $\beta$ that balances latent channel capacity and independence constraints with reconstruction accuracy, as formulated below:
\begin{equation}
\mathcal{L}_{\beta-VAE}(\mathbf{x}) = \mathbb{E}_{q(\mathbf{z}|\mathbf{x})}\left[\log p(\mathbf{x}|\mathbf{z})\right] - \beta D_{KL}\left(q(\mathbf{z}|\mathbf{x})||p(\mathbf{z})\right) 
\end{equation}
where the observation of an instance $\mathbf{x}$ is generated from the latent variable $\mathbf{z}$. \citet{DBLP:journals/corr/abs-1804-03599} discussed the disentangling in $\beta$-VAE from an information bottleneck perspective. When $\beta=1$, it is the same as VAE. When $\beta>1$, it applies a stronger constraint on the latent bottleneck and limits the representation capacity of $\mathbf{z}$. 
It has been demonstrated that $\beta$-VAE with appropriately tuned $\beta > 1$ qualitatively outperforms VAE ($\beta = 1$) \cite{DBLP:journals/corr/abs-1804-03599,DBLP:conf/iclr/HigginsMPBGBML17}. 


Multi-view representation learning based on $\beta$-VAE has been investigated in recent years to learn disentangled representations for multi-view data (e.g., DMVAE \cite{DBLP:conf/dagm/DaunhawerSMV20}, MEI \cite{DBLP:conf/mm/GuoHKH19}, Multi-VAE \cite{DBLP:conf/iccv/Xu0TP0Z021}, and TCM \cite{DBLP:journals/pami/HanZFZ23}). However, none of them can solve the aforementioned three issues holistically. In addition, $\beta$-VAE has a trade-off between reconstruction quality and disentanglement capacity. In this work, we propose a theoretical assessment bound capped on $\beta$-VAE loss to control disentanglement capacity. What's more, we design an intuitive discriminator to disentangle fusion representation and individuation representation. The details of the proposed model will be clear shortly.

\section{Proposed Model}
As aforementioned, we propose a dPoE for online multi-view anomaly detection. Prior to introducing the details of the model, we first clarify the problem, the intuition of our solutions, the model architecture, and the roadmap to our model.

\myparagraph{Problem Statement (\textit{Online Multi-view Anomaly Detection})}\\
Given a set of multi-view data with $m$ views $\mathcal{D}=\{\mathbf{X}^1,...,\mathbf{X}^m\}$, the data $\mathcal{D}$ are used to train a model (e.g., our dPoE) for an online multi-view anomaly detection task. The model is then employed to detect anomalies for a single data instance or a batch of data instances after deployment, where the test data instance(s) may originate from $\mathcal{D}$ or a newly collected dataset $\mathcal{\widetilde{D}}$.

\myparagraph{Intuition of Our Solutions} Building upon the existing literature \cite{DBLP:conf/icdm/GaoFTPH11,DBLP:conf/cikm/AlvarezYKI13,DBLP:conf/ijcai/ZhaoF15a,DBLP:conf/aaai/Li0DZF18,DBLP:conf/aaai/GuoZ18}, our core intuition is that data instances typically form groups or clusters. Consequently, we regard data instances that \textit{do not} belong to any clusters as anomalous instances. We formally define this intuition using probability as follows:
\begin{itemize}[leftmargin=*]
\item We denote the probability distribution of hidden clusters in the data as $p(\mathbf{\pi})=p(\pi_1)p(\pi_2)\cdots p(\pi_k)$.
\item We define the probability of an instance belonging to a cluster in each view as $p(\pi_i|\mathbf{x}^v)$. We integrate the probabilities across all views as $p(\pi_i|\{\mathbf{x}^v\})$. A lower $p(\mathbf{\pi}|\{\mathbf{x}^v\})$ value across all clusters indicates that the instance $x$ is more likely to be anomalous.
\end{itemize}

\begin{figure*}[!htbp]
    \centering
    \includegraphics[width=1.0\linewidth]{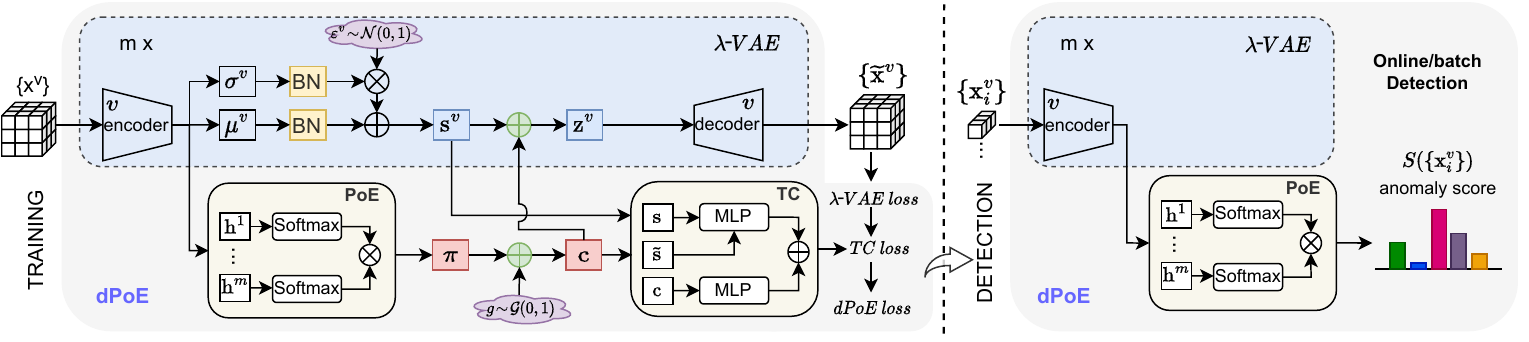}
    \caption{\small An overview of the proposed dPoE. It consists of multi-view VAE (i.e., $\lambda$-VAE), Product-of-Experts (PoE), and Total Correlation (TC) discriminator. All components are wrapped up by dPoE loss which is a joint loss of $\lambda$-VAE loss and TC loss. TRAINING (\textit{left}): The dPoE model and the TC discriminator are updated using the dPoE loss and the TC loss respectively in an alternative optimization scheme. DETECTION (\textit{right}): The trained model is then deployed for online (or batch) detection by using the encoder and PoE module. [Best viewed in color]}
    \label{framework}
\end{figure*}

\myparagraph{Architecture} Figure~\ref{framework} illustrates the architecture of our dPoE, comprising three novel components: (i) \textit{Multi-view Variational Autoencoder} (i.e., $\lambda$-VAE), which models multi-view data using a latent view-common representation $\mathbf{c}$ and view-specific representations $\{\mathbf{s}^1,...,\mathbf{s}^m\}$ in a generative approach, (ii) \textit{Product-of-Experts} (PoE), which integrates multiple ``experts'' to make decisions on data, and (iii) \textit{Total Correlation discriminator} (TC), which disentangles view-common representation from view-specific representations.

\myparagraph{Roadmap to Our Model} Having defined the problem, the intuition, and the architecture, our primary objectives for designing dPoE are (1) \textit{modeling multi-view data} using $\lambda$-VAE (\textbf{Section~\ref{sec3.1}}), (2) \textit{modeling anomaly detection} with PoE (\textbf{Section~\ref{sec3.2}}), (3) \textit{learning disentangled representations} using TC (\textbf{Section~\ref{sec3.3}}), and finally (4) \textit{online detection} with dPoE (\textbf{Section~\ref{sec3.4}}). We now elaborate on our solutions to each component.

\subsection{Modeling Multi-view Data}
\label{sec3.1}
\myparagraph{Generative Process}
In this work, we propose learning fusion representation (i.e., view-common representation $\mathbf{c}$) and individuation representation (i.e., view-specific representation $\{\mathbf{s}^1,...,\mathbf{s}^m\}$) together. We then concatenate $\mathbf{c}$ and $\{\mathbf{s}^1,...,\mathbf{s}^m\}$ as latent representation $\{\mathbf{z}^1,...,\mathbf{z}^m\}$ to reconstruct the input multi-view data $\{\mathbf{x}^1,...,\mathbf{x}^m\}$ in a generative manner, formulated as:
\begin{equation}
    \mathbf{\widetilde{x}}^v=Decoder(\mathbf{z}^v)
\end{equation}
where $\mathbf{z}^v=[\mathbf{c},\mathbf{s}^v]$, and $v=1,...,m$. 

Similar to VAE, we consider the following generative process:
\begin{equation}
    p(\mathbf{x}^v,\mathbf{s}^v,\mathbf{c}) = p(\mathbf{x}^v|\mathbf{s}^v,\mathbf{c})p(\mathbf{s}^v,\mathbf{c}) = p(\mathbf{x}^v|\mathbf{s}^v,\mathbf{c})p(\mathbf{s}^v)p(\mathbf{c})
\end{equation}
where $p(\mathbf{s}^v,\mathbf{c})=p(\mathbf{s}^v)p(\mathbf{c})$, assuming the view-common and view-specific variables are conditionally independent.

\myparagraph{Variational Lower Bound}
Given the generative process above, using Jensen's inequality, the log-likelihood of the given data instances can be defined as:
\begin{equation}
\begin{aligned}
    \sum_{v=1}^{m}\log p(\mathbf{x}^v) & = \sum_{v=1}^{m}\log \int_{\mathbf{s}^v}\sum_{\mathbf{c}}p(\mathbf{x}^v,\mathbf{s}^v,\mathbf{c})d\mathbf{s}^v \\
    & \ge \sum_{v=1}^{m}\mathbb{E}_{q(\mathbf{s}^v,\mathbf{c}|\mathbf{x}^v)}\left[\log \frac{p(\mathbf{x}^v,\mathbf{s}^v,\mathbf{c})}{q(\mathbf{s}^v,\mathbf{c}|\mathbf{x}^v)}\right] \\
    & = \sum_{v=1}^{m}\mathcal{L}_{ELBO}(\mathbf{x}^v)
\end{aligned}
\end{equation}
where $\mathcal{L}_{ELBO}$ is the Evidence Lower BOund (ELBO), and $q(\mathbf{s}^v,\mathbf{c}|\mathbf{x}^v)$ is the variational posterior that approximates the true posterior $p(\mathbf{s}^v,\mathbf{c}|\mathbf{x}^v)$. In the family of variational inference, maximizing the likelihood is equivalent to maximizing the ELBO. We further apply the mean-field approximation to factorize $q(\mathbf{s}^v,\mathbf{c}|\mathbf{x}^v)$ as $q(\mathbf{s}^v,\mathbf{c}|\mathbf{x}^v)=q(\mathbf{s}^v|\mathbf{x}^v)q(\mathbf{c}|\mathbf{x}^v)$. Then, the ELBO of each view is formulated as:
\begin{equation}\label{elbo}
\begin{aligned}
    \mathcal{L}_{ELBO}(\mathbf{x}^v) &= \mathbb{E}_{q(\mathbf{s}^v,\mathbf{c}|\mathbf{x}^v)}\left[\log p(\mathbf{x}^v|\mathbf{s}^v,\mathbf{c})\right] \\ 
    &- D_{KL}(q(\mathbf{s}^v|\mathbf{x}^v)||p(\mathbf{s}^v)) - D_{KL}(q(\mathbf{c}|\{\mathbf{x}^v\})||p(\mathbf{c}))
\end{aligned}
\end{equation}
where $D_{KL}(\cdot)$ denote the KL divergence of two distributions.

To elaborate, we aim at disentangled representation learning. As mentioned in Section~\ref{beta_vae}, $\beta$-VAE uses an adjustable hyperparameter to tackle disentangled representation learning. However, $\beta$-VAE has a trade-off between reconstruction quality and disentanglement capacity. As a result, $\beta$-VAE may suffer from limited disentanglement capability. To address this issue, we devise a theoretical bound on the ELBO to control disentanglement capacities for both view-common and view-specific representations, formulated as follows (which we call $\lambda$-VAE):
\begin{equation}
\begin{aligned}
    \mathcal{L}_{\lambda-VAE}(\mathbf{x}^v) &= \mathbb{E}_{q(\mathbf{s}^v,\mathbf{c}|\mathbf{x}^v)}\left[\log p(\mathbf{x}^v|\mathbf{s}^v,\mathbf{c})\right] \\ 
    &- \lambda \|D_{KL}(q(\mathbf{s}^v|\mathbf{x}^v)||p(\mathbf{s}^v))-C_{\mathbf{s}^v}\| \\
    &- \lambda \|D_{KL}(q(\mathbf{c}|\{\mathbf{x}^v\})||p(\mathbf{c}))-C_{\mathbf{c}}\|
\end{aligned}
\end{equation}
where $C_{\mathbf{s}^v}$ and $C_{\mathbf{c}}$ are the bounded capacities on the view-specific representation and view-common representation, respectively. We next derive a set of theoretical bound results ($C_{\mathbf{s}^v}=\frac{d^v}{2}$ and $C_\mathbf{c}=\log K$, where $d^v$ is the dimension of view-specific representation $\mathbf{s}^v$, and $K$ is the number of clusters hidden in the data\footnote{The $K$ can be a user-given prior or evaluated by an off-the-shelf method, e.g., \cite{tibshirani2001estimating}.}).

\textbf{Derivation of \underline{$C_{\mathbf{s}^v}=\frac{d^v}{2}$}.}
For view-specific representations, we propose modeling them using a Gaussian distribution $\mathcal{N}$. In practice, we infer $\mathbf{s}^v$ by drawing from a posterior distribution according to the reparameterization trick, formulated as:
\begin{equation}
    q(\mathbf{s}^v|\mathbf{x}^v)=\mathcal{N}(\mu^v,(\sigma^v)^2)=\mu^v+\epsilon^v\sigma^v
\end{equation}
where $[\mu^v,\sigma^v]=Encoder(\mathbf{x}^v)$, and $\epsilon^v \sim \mathcal{N}(0,1)$.

The KL divergence between the view-specific posterior and its prior is formulated as:
\begin{equation}
    D_{KL}(q(\mathbf{s}^v|\mathbf{x}^v)||p(\mathbf{s}^v)) = \frac{1}{2}\sum_{i=1}^{d^v}((\mu_i^v)^2+(\sigma_i^v)^2-\log (\sigma_i^v)^2-1)
\end{equation}
where $d^v$ is the dimension of the representation $\mathbf{s}^v$.
The expectation of the KL can be noted as:
\begin{equation}
\begin{aligned}
    \mathbb{E}[D_{KL}(q(\mathbf{s}^v|\mathbf{x}^v)||p(\mathbf{s}^v)) ] & = \frac{1}{2}\sum_{i=1}^d(Var[\mu_i^v] + \mathbb{E}^2[\mu_i^v] \\
    & + \mathbb{E}[(\sigma_i^v)^2] - \mathbb{E}[\log (\sigma_i^v)^2]-1) \\
    & \ge \frac{1}{2}\sum_{i=1}^d(Var[\mu_i^v] + \mathbb{E}^2[\mu_i^v])
\end{aligned}
\end{equation}
where $Var[\mu_i^v]$ denotes the variance of $\mu_i^v$. $\mathbb{E}[(\sigma_i^v)^2] - \mathbb{E}[\log (\sigma_i^v)^2]\! \ge \! 1$ due to a basic property $e^x-x \ge 1$. Since the prior is drawn from a standard Gaussian distribution $\mathcal{N}(0,1)$, it is expected that $Var[\mu_i^v]=1$ and $\mathbb{E}^2[\mu_i^v]=0$. Then, we derive a lower bound of the KL divergence about view-specific representation as below:
\begin{equation}
    D_{KL}(q(\mathbf{s}^v|\mathbf{x}^v)||p(\mathbf{s}^v)) \ge \frac{d^v}{2}.
\end{equation}

Thus, we set $C_\mathbf{s}^v=\frac{d^v}{2}$. In addition, the family of VAE often suffers from a well-known problem called the \textit{posterior collapse} or the \textit{KL vanishing} problem \cite{DBLP:conf/iclr/LucasTGN19}. Inspired by \cite{DBLP:conf/acl/ZhuBLMLW20}, we plug a Batch Normalization (BN) layer into $\lambda$-VAE.

\textbf{Derivation of \underline{$C_\mathbf{c}=\log K$}.}
Since our motivation is to find cluster information hidden in the data, we propose to model view-common representation $\mathbf{c}$ as a one-hot representation using a categorical distribution such that the learned representation can model cluster probability information. As shown in \cite{DBLP:conf/iccv/Xu0TP0Z021}, when $p(\mathbf{c})$ is a uniform categorical distribution, the KL divergence about the view-common posterior and its prior is bounded as:
\begin{equation}
\begin{aligned}
    D_{KL}(q(\mathbf{c}|\{\mathbf{x}^v\})||p(\mathbf{c})) &=\sum_{i=1}^{K}q(\mathbf{c}_i|\{\mathbf{x}^v\})\log \frac{q(\mathbf{c}_i|\{\mathbf{x}^v\})}{p(\mathbf{c}_i)} \\
    &= \sum_{i=1}^{K}q(\mathbf{c}_i|\{\mathbf{x}^v\})\log \frac{q(\mathbf{c}_i|\{\mathbf{x}^v\})}{1/K} \\
    &= -H(q(\mathbf{c}|\{\mathbf{x}^v\})) + \log K \le \log K
\end{aligned}
\end{equation}
where $H(\cdot)$ is the entropy. So, we set $C_\mathbf{c}=\log K$. In Section~\ref{sec3.2}, we will discuss how to infer the posterior $p(\mathbf{c}|\{\mathbf{x}^v\})$.

\subsection{Modeling Product-of-Experts}\label{sec3.2}
\myparagraph{Inference of PoE}
The first obstacle to modeling a PoE with $m$ experts is specifying the joint decisions (in a probability manner). We denote the decisions $\mathbf{\pi}$ on $\mathbf{x}^v$ by the expert $E_v$ as probabilities $p(\pi|\mathbf{e}^v)$ and assume that each expert makes decisions independently. Then, the joint probability can be formulated as:
\begin{equation}\label{for_poe}
\begin{aligned}
    p(\mathbf{\pi}|\mathbf{e}^1,\mathbf{e}^2,...,\mathbf{e}^m) &= \frac{p(\mathbf{e}^1,\mathbf{e}^2,...,\mathbf{e}_m|\mathbf{\pi})p(\mathbf{\pi})}{p(\mathbf{e}^1,\mathbf{e}^2,...,\mathbf{e}^m)} \\
    &= \frac{p(\mathbf{\pi})}{p(\mathbf{e}^1,\mathbf{e}^2,...,\mathbf{e}^m)}\prod_{i=1}^{m}p(\mathbf{e}^i|\mathbf{\pi}) \\
    &= \frac{p(\mathbf{\pi})}{p(\mathbf{e}^1,\mathbf{e}^2,...,\mathbf{e}^m)}\prod_{i=1}^{m}\frac{p(\mathbf{\pi}|\mathbf{e}^i)p(\mathbf{e}^i)}{p(\mathbf{\pi})} \\
    & \propto \frac{\prod_{i=1}^{m}p(\mathbf{\pi}|\mathbf{e}^i)}{\prod_{i=1}^{m-1}p(\mathbf{\pi})}.
\end{aligned}
\end{equation}

That is, the joint probability is a product of all individual probabilities with an additional quotient by the prior $p(\mathbf{\pi})$. If we assume that the true distribution for each individual factor $p(\mathbf{\pi}|\mathbf{e}^i)$ is properly contained in the family of its variational counterpart $\widetilde{q}(\mathbf{\pi}|\mathbf{e}^i)$, then Eq.~\eqref{for_poe} corrects to $\frac{\prod_{i=1}^{m}\widetilde{q}(\mathbf{\pi}|\mathbf{e}^i)}{\prod_{i=1}^{m-1}p(\mathbf{\pi})}$ as shown in \cite{DBLP:conf/nips/WuG18}. To avoid the quotient term, we approximate $p(\mathbf{\pi}|\mathbf{e}^i)$ with $\widetilde{q}(\mathbf{\pi}|\mathbf{e}^i) := q(\mathbf{\pi}|\mathbf{e}^i)p(\mathbf{\pi})$, where $q(\mathbf{\pi}|\mathbf{e}^i)$ is the underlying inference network. Eq.~\eqref{for_poe} is then simplified as follows:
\begin{equation}\label{poe}
\begin{aligned}
    p(\mathbf{\pi}|\{\mathbf{e}^v\}) & \propto \frac{\prod_{i=1}^{m}p(\mathbf{\pi}|\mathbf{e}^i)}{\prod_{i=1}^{m-1}p(\mathbf{\pi})} \\
    & \approx \frac{\prod_{i=1}^{m}[q(\mathbf{\pi}|\mathbf{e}^i)p(\mathbf{\pi})]}{\prod_{i=1}^{m-1}p(\mathbf{\pi})} \\
    & = p(\mathbf{\pi})\prod_{i=1}^{m}q(\mathbf{\pi}|\mathbf{e}^i).
\end{aligned}
\end{equation}

For the approximate factor $q(\mathbf{\pi}|\mathbf{e}^i)$, we infer it as 
\begin{equation}
    q(\mathbf{\pi}|\mathbf{e}^i) = Softmax(\mathbf{h}^i)
\end{equation}
where $\mathbf{h}^i$ is a hidden representation from an encoder, formulated as 
\begin{equation}
    \mathbf{h}^i = Encoder(\mathbf{x}^i).
\end{equation}

From Eq.~\eqref{poe}, we can see it is exactly a kind of PoE. Each expert is implemented with a Softmax function. Therefore, based on Eq.~\eqref{poe}, we define the joint decision probability for $\{\mathbf{x}^v\}$ by PoE as
\begin{equation}\label{poe_decision}
    p(\mathbf{\pi}_k|\{\mathbf{x}^v\}) = \frac{p(\mathbf{\pi}_k|\{\mathbf{e}^v\})}{\sum_{j=1}^{K}p(\mathbf{\pi}_j|\{\mathbf{e}^v\})} = \frac{\prod_{i=1}^{m}q(\mathbf{\pi}_k|\mathbf{e}^i)}{\sum_{j=1}^{K}[\prod_{i=1}^{m}q(\mathbf{\pi}_j|\mathbf{e}^i)]}
\end{equation}
where $p(\mathbf{\pi}_k|\{\mathbf{x}^v\})$ denotes the probability of an instance belonging to the cluster $k$. It is worth noting that the prior factor $p(\mathbf{\pi})$ has been left out in Eq.~\eqref{poe_decision}. What's more, such a PoE can tackle more than two views and identify all three types of anomalies as the output is a product of the decisions from $m$ experts. 

\myparagraph{Inference of $\mathbf{c}$}
The joint decisions are then considered to learn view-common representation $\mathbf{c}$. In practice, we approximate the true posterior $p(\mathbf{c}|\{\mathbf{x}^v\})$ as 
\begin{equation}
    q(\mathbf{c}|\{\mathbf{x}^v\}) = \prod_{k=1}^{K}q(\mathbf{c}_k|\{\mathbf{x}^v\})
\end{equation}
where $q(\mathbf{c}_k|\{\mathbf{x}^v\})$ is drawn from a Gumbel distribution $\mathcal{G}$ with the Gumbel-Softmax reparameterization trick \cite{DBLP:conf/iclr/JangGP17} as below
\begin{equation}
    q(\mathbf{c}_k|\{\mathbf{x}^v\}) = \frac{\exp((\log \mathbf{\pi}_k + g_k)/\tau)}{\sum_{i=1}^{K}\exp((\log \mathbf{\pi}_i + g_i)/\tau)} 
\end{equation}
where $g_k\sim \mathcal{G}(0,1)$, and $\tau$ is a temperature parameter allowing the distribution to concentrate its mass around the vertices. However, the above learned view-common and view-specific representations would be entangled with each other as they are learned independently in a purely self-supervised manner. We next present a discriminator to disentangle them.

\subsection{Learning Disentangled Representations}\label{sec3.3}
As aforementioned, a key challenge to learning view-common representation and view-specific representation simultaneously is disentangling them. To this end, we propose to push them away from each other by minimizing their KL divergence $D_{KL}(p(\mathbf{c})||p(\mathbf{s}^v))$. However, this term is intractable since both $p(\mathbf{c})$ and $p(\mathbf{s}^v)$ involve mixtures with a large number of components, and the direct Monte Carlo estimate requires a pass through the entire dataset for evaluation. $D_{KL}(p(\mathbf{c})||p(\mathbf{s}^v))$ is also known as Total Correlation (TC) \cite{DBLP:journals/ibmrd/Watanabe60}. According to \cite{DBLP:conf/icml/KimM18,DBLP:conf/dagm/DaunhawerSMV20}, we estimate TC using the density-ratio trick \cite{DBLP:journals/tit/NguyenWJ10} with a discriminator $\mathfrak{D}$ as follows:
\begin{equation}
\begin{aligned}
    TC(\mathbf{c},\mathbf{s}^v) &= D_{KL}(p(\mathbf{c})||p(\mathbf{s}^v)) = D_{KL}(q(\mathbf{c},\mathbf{s}^v)||q(\mathbf{c})q(\mathbf{s}^v)) \\
    &=  \mathbb{E}_{q(\mathbf{c},\mathbf{s}^v)}\left[\log \frac{q(\mathbf{c},\mathbf{s}^v)}{q(\mathbf{c})q(\mathbf{s}^v)}\right] \\
    &\approx  \mathbb{E}_{q(\mathbf{c},\mathbf{s}^v)}\left[\log \frac{\mathfrak{D}(\mathbf{c},\mathbf{s}^v)}{1-\mathfrak{D}(\mathbf{c},\mathbf{s}^v)}\right] \\
    & =: \mathcal{L}_{TC}(\mathbf{c},\mathbf{s}^v).
\end{aligned}
\end{equation}

\myparagraph{Objection Function}
We saddle $\lambda$-VAE with the TC discriminator. Formally, we have the total loss function of our dPoE:
\begin{equation}\label{dpoe_loss}
    \mathcal{L}_{dPoE}(\{\mathbf{x}^v\}) = \sum_{i=1}^{m}\mathcal{L}_{\lambda-VAE}(\mathbf{x}^i) - \sum_{i=1}^{m}\gamma \mathcal{L}_{TC}(\mathbf{c},\mathbf{s}^i)
\end{equation}
where $\gamma$ is a trade-off parameter. We train the TC discriminator and dPoE jointly. In particular, the dPoE parameters are updated using the objective function in Eq.~\eqref{dpoe_loss}. The discriminator is trained to distinguish $\mathbf{c}$ from $\mathbf{s}^v$ and a randomly sampled $\mathbf{\widetilde{s}}^v$ from the same batch by minimizing $\log \mathfrak{D}(\mathbf{c},\mathbf{s}^v) + \log (1-\mathfrak{D}(\mathbf{c},\mathbf{\widetilde{s}}^v))$. Therefore, we adopt an alternative optimization scheme, which alternatively updates the TC discriminator and dPoE. We employ Stochastic Gradient Variational Bayes (SGVB) \cite{DBLP:journals/corr/KingmaW13} estimator to train the model in an end-to-end manner with alternative gradient descent.

\subsection{Online Detection}\label{sec3.4}
We now discuss why our dPoE supports online detection. Given an instance with its data $\{\mathbf{x}^v\}$, we collect the decisions of all experts and define the anomaly score of this instance based on Eq.~\eqref{poe_decision} as 
\begin{equation}\label{anomaly_score}
    \mathcal{S}(\{\mathbf{x}^v\}) = \arg \max_k\left(1- \frac{\prod_{i=1}^{m}q(\mathbf{\pi}_k|\mathbf{e}^i)}{\sum_{j=1}^{K}[\prod_{i=1}^{m}q(\mathbf{\pi}_j|\mathbf{e}^i)]}\right).
\end{equation}

\myparagraph{Remarks}
Since the decision of each expert $q(\mathbf{\pi}_k|\mathbf{e}^i)$ in Eq.~\eqref{anomaly_score} can be directly inferred after model deployment, it is clear that our dPoE supports online detection. As the final decision on a test instance is a product of multiple experts, dPoE can tackle three types of anomalies holistically. For \textsc{Type-I}, every expert would give a low decision score as they did not see such kind of instances. For \textsc{Type-II}, different experts would make inconsistent decisions. So, the final decision scores would be low. For \textsc{Type-III}, some experts would give low decision scores, and the decisions of all experts would also be inconsistent. In addition, it is worth noting that the computational complexity of our model is $\mathcal{O}(mNK^2)$. We omit the constants $m$ and $K$ and finalize that the computational complexity of optimizing dPoE is linear to the data size $N$. Therefore, the time complexity of online detection is $\mathcal{O}(1)$.
\section{Experiments}
\subsection{Experimental Setup}
\myparagraph{Datasets}
We experiment on six real-world multi-view datasets. All the used datasets are publicly available. The statistics of each dataset are shown in Table~\ref{dataset}. In this table, MNIST~\cite{DBLP:journals/pieee/LeCunBBH98} is a classic handwritten digital image dataset (0-9). Fashion-MNIST~\cite{DBLP:journals/corr/abs-1708-07747} (or shortly Fashion) is a novel image dataset of ten categories of fashion products. COIL-20~\cite{nene1996columbia} is another widely-used image dataset of twenty objects. Following~\cite{DBLP:conf/iccv/Xu0TP0Z021}, we construct a multi-view version for these three datasets. Specifically, different people writing for the same digit is treated as different views. Different fashions designing for the same category of products are treated as different views. The same object in different poses refers to it has multiple views. Pascal-Sentences~\cite{DBLP:conf/naacl/RashtchianYHH10} (or simply Pascal-S) is a multi-modality dataset that consists of image and text, where each image is annotated with five English sentences. We treat each modality as a view. Fox-News~\cite{DBLP:conf/cikm/QianZ14} and CNN-News~\cite{DBLP:conf/cikm/QianZ14} (Fox-N and CNN-N in short) are CNN and FOX web news data, respectively. Each news has an image view and a text view. For text view, we use the text in its abstract body.
\begin{table}[!htbp]
    \small
    \centering
    \caption{\small Data statistics of the benchmark datasets.}
    \resizebox{0.475\textwidth}{!}{
    \begin{tabular}{c|cccc}
    \toprule
    \addlinespace[2pt]
        \textsc{Dataset} & \#\textsc{Instances} & \#\textsc{Views} & \#\textsc{Classes} & \textsc{Feature Type} \\
    \addlinespace[2pt]
    \midrule
    \addlinespace[2pt]
        MNIST~\cite{DBLP:journals/pieee/LeCunBBH98} & 70,000 & 2 & 10 & \multirow{3}{*}{Image} \\
        Fashion~\cite{DBLP:journals/corr/abs-1708-07747} & 10,000 & 3 & 10 \\
        COIL-20~\cite{nene1996columbia} & 1,440 & 3 & 20 \\
    \addlinespace[2pt]
    \midrule
    \addlinespace[2pt]
        Pascal-S~\cite{DBLP:conf/naacl/RashtchianYHH10} & 1,000 & 2 &  20 & \multirow{3}{*}{Image\&Text}\\
        FOX-N~\cite{DBLP:conf/cikm/QianZ14} & 1,523 & 2 & 4 \\
        CNN-N~\cite{DBLP:conf/cikm/QianZ14} & 2,107 & 2 & 7\\
    \addlinespace[2pt]
    \bottomrule
    \end{tabular}}
    \label{dataset}
\end{table}

\myparagraph{Baselines}
We compare dPoE against the following nine multi-view anomaly detection methods: HOAD~\cite{DBLP:conf/icdm/GaoFTPH11}, AP~\cite{DBLP:conf/cikm/AlvarezYKI13}, MLRA~\cite{DBLP:conf/sdm/LiSF15}, LDSR~\cite{DBLP:conf/aaai/Li0DZF18}, CL~\cite{DBLP:conf/aaai/GuoZ18}, HBM~\cite{DBLP:conf/ijcai/WangL20}, SRLSP~\cite{DBLP:journals/tkdd/WangCLFZZ23}, MODDIS~\cite{DBLP:conf/icdm/JiHHWXSL19}, and NCMOD~\cite{DBLP:conf/aaai/Cheng0L21}. As shown in Table~\ref{related_work_summary}, MODDIS and NCMOD are two deep learning-based methods. The other seven baselines are non-deep learning methods. We note that HOAD requires an appropriate hyperparameter to control how closely different views of the same instance are embedded together. We ran HOAD with a grid search of settings $\{0.1, 1, 10, 100\}$ for each dataset. For AP, we follow the settings of MLRA, LDSR and CL to utilize the $l_2$ distance with HSIC. For the other baselines, we ran their systems with the settings as they suggested. In addition, the architectures of HOAD, AP, MLRA, and CL are tailored for two-view data. To evaluate them on more than two-view data, we follow the settings of SRLSP and MODDIS that calculate anomaly scores in each pair of views and then take average values over the scores as the final anomaly scores. 

\begin{table*}[!htbp]
   \small
    \centering
    \caption{\small Anomaly detection performance in terms of AUC on three image-based multi-view datasets. The symbol ``--'' denotes that such a method ran out of memory. So we cannot collect its outputs. We ran those methods on a Windows OS with Intel Xeon CPU and 32 GB RAM.}
    \vspace{-0.1cm}
    \resizebox{\textwidth}{!}{
    \begin{tabular}{lcccccccccccccccc}
    \toprule
    \addlinespace[2pt]
        \multirow{2}{*}{\textsc{Method}} & \multicolumn{3}{c}{\textsc{Type-I}} & \multicolumn{3}{c}{\textsc{Type-II}} & \multicolumn{3}{c}{\textsc{Type-III}} & \multicolumn{3}{c}{\textsc{Type-Mix}}  \\
        \cmidrule(lr){2-4}\cmidrule(lr){5-7}\cmidrule(lr){8-10}\cmidrule(lr){11-13}
        & MNIST & Fashion & COIL-20 & MNIST & Fashion & COIL-20 & MNIST & Fashion & COIL-20 & MNIST & Fashion & COIL-20 \\
    \addlinespace[2pt]
    \midrule
    \addlinespace[2pt]
        HOAD~\cite{DBLP:conf/icdm/GaoFTPH11} &   --   &   0.7651    &   0.3461    &   --   &   0.5108    &   0.6007    &   --   &   0.6635    &   0.5262    &   --   &   0.6715    &   0.5869  \\
        AP~\cite{DBLP:conf/cikm/AlvarezYKI13} &   --   &   0.4620    &   0.3131    &   --   &   0.5119    &   0.6361    &   --   &   0.6317    &   0.4552    &   --   &   0.6072    &   0.5583   \\
        MLRA~\cite{DBLP:conf/sdm/LiSF15} & 0.7306 &   0.7810    &   0.8426    & 0.6527 &   0.6975    &   0.6908    & 0.8696 &   0.8928    &   0.9035    & 0.8012 &   0.8356    &   0.8643   \\
        LDSR~\cite{DBLP:conf/aaai/Li0DZF18} &   --   &   0.8345    &   0.9556    &   --   &   0.7998    &   0.9025    &   --   &   0.8072    &   0.9211    &   --   &   0.8112    &   0.9025    \\
        CL~\cite{DBLP:conf/aaai/GuoZ18} &   --   &   0.4630    &   0.5916    &   --   &   0.6221    &   0.7976    &   --   &   0.5360    &   0.8374    &   --   &   0.5506    &   0.8276    \\
        HBM~\cite{DBLP:conf/ijcai/WangL20} & 0.7083 &   0.6720    &   0.6217    & 0.6338 &   0.7411    &   0.6935    & 0.6834 &   0.7005    &   0.6780    & 0.6710 &   0.7178    &   0.6702    \\
        SRLSP~\cite{DBLP:journals/tkdd/WangCLFZZ23} & 0.9612 &   0.9524    &   0.9623    & 0.7743 &   0.8108    &   0.9193    & 0.9321 &   0.9587    &   0.9624    & 0.9125 &   0.9045    &   0.9337    \\
        MODDIS~\cite{DBLP:conf/icdm/JiHHWXSL19} & 0.8876 &   0.8687    &   0.9187    & 0.8332 &   0.7834    &   0.8262    & 0.8820 &   0.7923    &   0.8760    & 0.8576 &   0.8103    &   0.8675    \\
        NCMOD~\cite{DBLP:conf/aaai/Cheng0L21} & 0.9049 &   0.8921    &   0.8932    & 0.8794 &   0.8360    &   0.8593    & 0.8787 &   0.8834    &   0.9051    & 0.8796 &   0.8554    &   0.8580    \\
        dPoE (ours) & \textbf{0.9940} &   \textbf{0.9763 }   &  \textbf{0.9970}   & \textbf{0.9255} &   \textbf{0.8920}    &   \textbf{0.9427}    & \textbf{0.9442} &   \textbf{0.9704}    &  \textbf{0.9896}   & \textbf{0.9512} &   \textbf{0.9374}    &   \textbf{0.9657}   \\
    \addlinespace[2pt]
    \bottomrule
    \end{tabular}}
    \label{table3}
\end{table*}

\begin{table*}[!htbp]
    \small
    \centering
    \caption{\small Anomaly detection performance in terms of AUC on three image-and-text based multi-view datasets. The results of MLRA are marked with ``--'' due to MLRA system requires that the feature dimension of each view should be equal. However, the datasets (Pascal-S, FOX-N, and CNN-N) do not meet such a condition.}
    \vspace{-0.1cm}
    \resizebox{\textwidth}{!}{
    \begin{tabular}{lcccccccccccccccc}
    \toprule
    \addlinespace[2pt]
        \multirow{2}{*}{\textsc{Method}} & \multicolumn{3}{c}{\textsc{Type-I}} & \multicolumn{3}{c}{\textsc{Type-II}} & \multicolumn{3}{c}{\textsc{Type-III}} & \multicolumn{3}{c}{\textsc{Type-Mix}}  \\
        \cmidrule(lr){2-4}\cmidrule(lr){5-7}\cmidrule(lr){8-10}\cmidrule(lr){11-13}
        & Pascal-S & FOX-N & CNN-N & Pascal-S & FOX-N & CNN-N & Pascal-S & FOX-N & CNN-N & Pascal-S & FOX-N & CNN-N \\
    \addlinespace[2pt]
    \midrule
    \addlinespace[2pt]
        HOAD~\cite{DBLP:conf/icdm/GaoFTPH11} &  0.4650  & 0.6320 & 0.6372 &  0.3976  & 0.5825 & 0.5306 &  0.4379  & 0.6017 & 0.5905 &  0.4573  & 0.5825 & 0.6013 \\
        AP~\cite{DBLP:conf/cikm/AlvarezYKI13} &  0.5169  & 0.5793 & 0.5575 &  0.4312  & 0.5893 & 0.6014 &  0.5134  & 0.5869 & 0.6074 &  0.4914  & 0.5870 & 0.5820 \\
        MLRA~\cite{DBLP:conf/sdm/LiSF15} &    --    &   --   &   --   &    --    &   --   &   --   &    --    &   --   &   --   &    --    &   --   &   --   \\
        LDSR~\cite{DBLP:conf/aaai/Li0DZF18} &  0.6125  & 0.7324 & 0.7098 &  0.6049  & 0.6156 & 0.6226 &  0.6112  & 0.6954 & 06773  &  0.6139  & 0.6990 & 0.6872 \\
        CL~\cite{DBLP:conf/aaai/GuoZ18}  &  0.5956  & 0.5321 & 0.5527 &  0.4930  & 0.5998 & 0.5879 &  0.5405  & 0.5780 & 0.5734 &  0.5629  & 0.5507 & 0.5765 \\
        HBM~\cite{DBLP:conf/ijcai/WangL20} &  0.5789  & 0.6571 & 0.6523 &  0.5891  & 0.6324 & 0.6017 &  0.5827  & 0.6400 & 0.6378 &  0.5824  & 0.6455 & 0.6577 \\
        SRLSP~\cite{DBLP:journals/tkdd/WangCLFZZ23} &  0.6395  & 0.7312 & 0.7078 &  0.6023  & 0.6514 & 0.6678 &  0.6079  & 0.7021 & 0.6995 &  0.6110  & 0.6829 & 0.6865 \\
        MODDIS~\cite{DBLP:conf/icdm/JiHHWXSL19}  &  0.6590  & 0.7366 & 0.7512 &  0.6029  & 0.6378 & 0.6998 &  0.6334  & 0.6954 & 0.7304 &  0.6445  & 0.7125 & 0.7305 \\
        NCMOD~\cite{DBLP:conf/aaai/Cheng0L21} &  0.6749  & 0.7512 & 0.7735 &  0.6321  & 0.6556 & 0.7034 &  0.6548  & 0.6939 & 0.7327 &  0.6523  & 0.6923 & 0.7579 \\
        dPoE (ours) &  \textbf{0.7023}  & \textbf{0.7926} & \textbf{0.8141} &  \textbf{0.6465}  & \textbf{0.7081} & \textbf{0.7820} &  \textbf{0.6672}  & \textbf{0.7524} & \textbf{0.8013} &  \textbf{0.6784}  & \textbf{0.7355} & \textbf{0.8024} \\
    \addlinespace[2pt]
    \bottomrule
    \end{tabular}}
    \label{table4}
\end{table*}

\myparagraph{Implementation Details}
As shown in Figure~\ref{framework}, the proposed dPoE consists of $\lambda$-VAE, PoE, and TC. Specifically, $\lambda$-VAE is a set of $m$ stacked encoder and decoder blocks. Each encoder consists of three convolutional (Conv) layers and a fully connected (Fc) layer\footnote{Note that, more modern encoders (e.g., Transformer encoder) can also be employed for dPoE in a plug-and-play manner. Encoder is not the main scope of this paper.}: Input-Conv$_{32}^4$-Conv$_{64}^4$-Conv$_{64}^4$-Fc$_{256}$. Decoders are symmetric with respect to the encoders. The activation function for $\lambda$-VAE is ReLU. The stride is set to 2, and the dimension of each view-specific representation is set to 10. PoE consists of $m$ experts with each being formulated by a Softmax function. TC discriminator consists of two mapping modules with three Fc layers (Fc$_{500}$-Fc$_{500}$-Fc$_{500}$), and one score module with three Fc layers (Fc$_{1000}$-Fc$_{1000}$-Fc$_{2}$). The activation function for TC is LeakyReLU. We use Adam with the learning rate of $1e^{-4}$ to jointly train dPoE and TC for 500 epochs. Throughout, $\lambda$ and $\gamma$ are set to 50 unless otherwise stated. For image data, we rescale each image to size 32x32. For text data, we use the pre-trained embedding GloVe.840B~\cite{DBLP:conf/emnlp/PenningtonSM14} to initialize word vectors, and then follow \cite{DBLP:conf/emnlp/Kim14} to feed the vectors into dPoE. We flatten the features in tensor into a vector as the inputs for those non-deep learning baselines.

\subsection{Experimental Results}
\myparagraph{Anomaly Detection Settings}
Following the above-mentioned baseline methods, we use \textit{generated anomalies} for the evaluation. We generate each type of anomalies as below. \textsc{Type-I}: We randomly select an instance, and perturb its feature in all views by random values \cite{DBLP:conf/ijcai/ZhaoF15a}. \textsc{Type-II}: We randomly sample two instances from two different classes and then swap their features in one view but not in the other view(s) \cite{DBLP:conf/icdm/GaoFTPH11}. \textsc{Type-III}: We randomly sample two instances from two different classes and then swap the feature vectors in one view, while perturbing the feature of the two instances with random values in the other view(s) \cite{DBLP:conf/aaai/Li0DZF18}. The proportion of the generated anomalies in each dataset is set to $10\%$. In addition, we follow the recent work \cite{DBLP:conf/icdm/JiHHWXSL19,DBLP:conf/aaai/Cheng0L21,DBLP:journals/tkdd/WangCLFZZ23} and generate a mixed type of anomalies (denote as \textsc{Type-Mix}), which is a mixture set of the above three types with each type taking $5\%$ proportion. 

We use AUC (area under ROC curve) as the evaluation metric to evaluate the performance of each method. 

\myparagraph{Overall Evaluation}
The evaluation results in terms of AUC on each dataset with each type of anomaly are shown in Tables~\ref{table3}-\ref{table4}. From the tables, we have the following observations:
\begin{itemize}[leftmargin=*]
    \item Our proposed dPoE model is markedly better than all baseline methods. Our dPoE achieves the best performance in terms of AUC values on all the evaluation datasets. The results clearly show the effectiveness of the proposed dPoE model.
    \item Among those non-deep learning methods (i.e., HOAD, AP, MLRA, LDSR, CL, HBM, SRLSP), SRLSP is a promising method, especially for those three image-based multi-view datasets. However, most of these non-deep learning methods are inferior to the deep learning-based methods (i.e., MODDIS, NCMOD, and our dPoE). One reason is that deep learning methods usually have powerful nonlinear data-fitting abilities. In addition, deep learning methods are more suitable for tackling large-scale and high-dimensional data. The evaluation datasets are in this case. 
    
    \item Our dPoE outperforms those two deep learning baselines (MODDIS, NCMOD) mainly due to that they fused multi-view data roughly and also did not consider the problem of fusion disentanglement. Besides, they aim at employing deep learning for data representations. None of them is an end-to-end deep learning-based multi-view anomaly detection framework.
    \item Anomaly detection on multi-modal datasets (i.e.,  Pascal-S, FOX-N, CNN-N) is more challenging than on single-modal multi-view datasets (i.e., MNIST, Fashion, COIL-20). Such a challenge tends to be related to the representation of the semantic information in text data. This indicates that anomaly detection on multi-modal data calls for new representation techniques.
\end{itemize}


\subsection{Parameter Sensitivity Analysis}
\begin{figure}[!htp]
    \centering
    \includegraphics[width=0.24\textwidth]{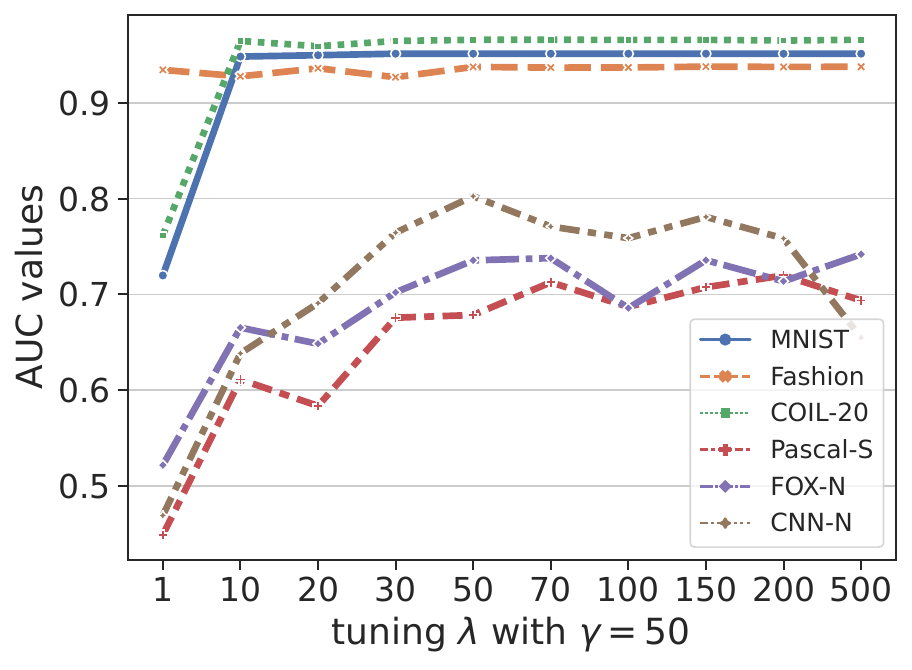}%
    \includegraphics[width=0.24\textwidth]{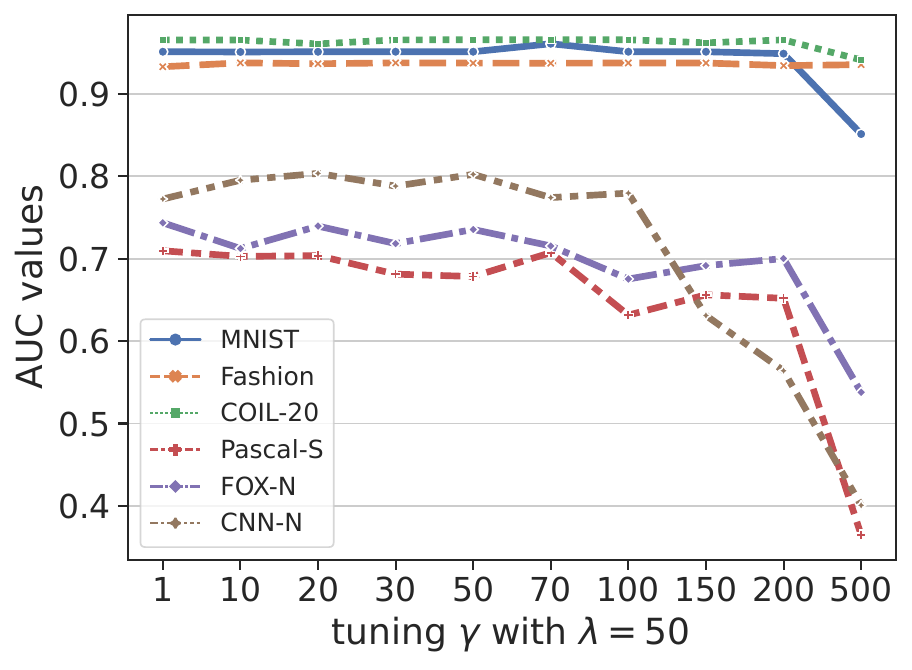}
    \vspace{-0.4cm}
    \caption{\small AUC curves against parameter $\lambda$ (\textit{left}) and $\gamma$ (\textit{right}).}
    \label{param_study}
    \vspace{-0.4cm}
\end{figure}

Since our dPoE includes two hyperparameters (i.e., VAE trade-off coefficient $\lambda$ and TC trade-off coefficient $\gamma$), we study the effect of each parameter. As shown in Figure~\ref{param_study}, we employ a grid search of settings $\{1, 10, 20, 30, 50, 70, 100, 150, 200, 500\}$. We fix one parameter and then tune another parameter. From the results, we can see that both two parameters have a large range of settings regarding high performance for our model. The model performance does not degrade markedly even when $\lambda$ is set to a large value. The reason is clear that we apply a capped bound on the model. A larger $\gamma$ would lead to significant performance decay due to that a larger $\gamma$ would bring more biases from TC discriminator to the model. In summary, the range $[20, 100]$ is optimal for dPoE.


\subsection{Ablation Study}
We conduct an ablation study to measure the performance gain by each of the four key components: view-common capacity $C_\mathbf{c}$ in Section~\ref{sec3.1}, view-specific capacity $C_{\mathbf{s}^v}$ in Section~\ref{sec3.1}, PoE in Section~\ref{sec3.2}, and TC in Section~\ref{sec3.3}. We remove each of them from dPoE and then evaluate each variant on six datasets with the anomaly type of \textsc{Type-Mix}. The evaluation results are shown in Table~\ref{ablation_study}. From the results, we can see that (1) every component plays an important role in the model, (2) view-common capacity is crucial for model performance, and (3) TC discriminator may have a side effect on the model performance in some cases.
\begin{table}[!htbp]
    \small
    \centering
    \caption{\small Ablation study on the variants of dPoE.}
    \vspace{-0.2cm}
    \resizebox{0.475\textwidth}{!}{
    \begin{tabular}{lcccccc}
    \toprule
    \addlinespace[2pt]
        \multirow{2}{*}{\textsc{Variants}} & \multicolumn{6}{c}{\textsc{Type-Mix}} \\
        \cmidrule(lr){2-7}
        & MNIST & Fashion & COIL-20 & Pascal-Sen & FOX-News & CNN-News \\
    \addlinespace[2pt]
    \midrule
    \addlinespace[2pt]
        w/o $C_\mathbf{c}$ & 0.5171 & 0.4710  & 0.3278  &  0.5031  & 0.5069 & 0.5534 \\
        w/o $C_{\mathbf{s}^v}$ & 0.8056 & 0.8499  & 0.9527  &  0.6012  & 0.6573 & 0.7621 \\
        w/o POE & 0.7335 & 0.7984  & 0.8724  &  0.5978  & 0.6275 & 0.6129 \\
        w/o TC & \textbf{0.9594} & 0.9005  & 0.9212  &  0.6320  & 0.6972 & 0.7845 \\
        dPoE & 0.9512 & \textbf{0.9374}  & \textbf{0.9657}  &  \textbf{0.6784}  & \textbf{0.7355} & \textbf{0.8024} \\
    \addlinespace[2pt]
    \bottomrule
    \end{tabular}}
    \label{ablation_study}
    \vspace{-0.2cm}
\end{table}

\section{Related Work}
Approaching to multi-view data has a rich history \cite{DBLP:conf/colt/BlumM98}. There are a large number of great relevant papers \cite{DBLP:journals/bigdatama/YangW18,DBLP:journals/tkde/LiYZ19,DBLP:journals/ijon/YanHMYY21}. The work on anomaly detection for multi-view data can date back to 2010s (\citet{DBLP:conf/icdm/GaoFTPH11}). They proposed a clustering-based method (called HOAD) to address \textsc{Type-II} anomalies, which refer to the ``inconsistent behavior'' in multi-view data. 
Following HOAD, AP (Affinity Propagation)~\cite{DBLP:conf/cikm/AlvarezYKI13}, PLVM (Probabilistic Latent Variable Models)~\cite{DBLP:conf/nips/IwataY16}, and CL (Collective Learning)~\cite{DBLP:conf/aaai/GuoZ18} were subsequently proposed to address \textsc{Type-II} anomaly detection. Among them, AP and CL are pair-wise constraint methods such that they are only suitable for two-view data. After that, there are some new research topics for multi-view anomaly detection \cite{tnnls/wang22}.

DMOD~\cite{DBLP:conf/ijcai/ZhaoF15a} is a pioneer work to address both \textsc{Type-I} and \textsc{Type-II} anomalies by representing multi-view data with latent coefficients and data-specific errors. DMOD was further extended to tackle more than two-view data by learning a consensus representation across different views \cite{DBLP:journals/tip/ZhaoLDF18}. MLRA (Multi-view Low-Rank)~\cite{DBLP:conf/sdm/LiSF15}, MLRA+ (an extension of MLRA)~\cite{DBLP:journals/tkdd/LiSF18}, and MuvAD (MUlti-View Anomaly Detection)~\cite{DBLP:conf/aaai/ShengZL019} are some follow-up work for both \textsc{Type-I} and \textsc{Type-II} anomaly detection. However, the architectures of MLRA, MLRA+, and MuvAD are designed for two-view data. 

\citet{DBLP:conf/aaai/Li0DZF18} raised the problem of detecting three types of anomalies simultaneously, and proposed a latent discriminant subspace representation (LDSR) method. IMVSAD (Inductive Multi-view Semi-Supervised Anomaly Detection)~\cite{DBLP:conf/icbk/WangFML19}, HBM (Hierarchical Bayesian Model)~\cite{DBLP:conf/ijcai/WangL20}, MBOD (Multi-view Bayesian Outlier Detector)~\cite{DBLP:conf/pakdd/WangZCLLXK21}, and SRLSP (Self-Representation method with Local Similarity Preserving)~\cite{DBLP:journals/tkdd/WangCLFZZ23} are most recent methods. However, all the above-mentioned methods are non-deep learning methods. CGAEs~\cite{DBLP:conf/ictai/WangLCZ21} is a multi-view autoencoder model for \textsc{Type-I-II} anomaly detection. FMOD (Fast Multi-view Outlier Detection)~\cite{DBLP:journals/tkdd/WangCLFZZ23}, MODDIS (Multi-view Outlier Detection in Deep Intact Space)~\cite{DBLP:conf/icdm/JiHHWXSL19}, and NCMOD (Neighborhood Consensus networks based Multi-view Outlier Detection)~\cite{DBLP:conf/aaai/Cheng0L21} are three relevant deep learning methods for \textsc{Type-I-II-III} anomaly detection. However, FMOD does not support online detection. MODDIS and NCMOD need to retain the hidden representation of the training data for supporting online detection, and none of them concern the problem of fusion disentanglement. 

In addition, \citet{DBLP:conf/cikm/WangSZWL18} investigated multi-view group anomaly detection, where individual instances might be normal but their collective pattern as a group is abnormal. MLRA+ \cite{DBLP:journals/tkdd/LiSF18} is also suited for multi-view group anomaly detection .





\section{Conclusion}
In this paper, we proposed a novel generative model (namely dPoE) for multi-view anomaly detection that considers model versatility, fusion disentanglement, and online detection. Specifically, we designed a Product-of-Experts layer and a new discriminator to learn disentangled view-common and view-specific representations. Besides, we devised a capacity bound to cap each representation. The experimental results using real-world datasets demonstrate the effectiveness of the proposed dPoE. Our future work includes: 1) integrating more advanced deep encoders into dPoE to enhance data representations, especially for tabular data, and 2) adapting dPoE to incomplete multi-view and semi-supervised settings.

\begin{acks}
{\small 
This work was supported in part by grants from the National Natural Science Foundation of China (no. 61976247), the Natural Science Foundation of Sichuan Province (no. 2022NSFSC0528), and Sichuan Science and Technology Program (no. 2022ZYD0113). Xiao Wu was supported by the Key R\&D Program of Guangxi Zhuang Autonomous Region (nos. AB22080038, AB22080039). Hongyang Chen was supported by the Key Research Project of Zhejiang Lab (no. 2022PI0AC01).}
\end{acks}

\bibliographystyle{ACM-Reference-Format}
\balance
\bibliography{sample-base}

\appendix
\appendix

\section{No Free Fusion Theorem}\label{proof}
We recall the ``no free fusion'' theorem (i.e., Theorem 1) in the main body of this paper as follows:

\myparagraph{Theorem 1 (No Free Fusion)} Let $\{\mathbf{x}^v$\} and $y$ be random variables with joint distribution $p(\{\mathbf{x}^v\},y)$. Let $\mathbf{z}^c$ and $\mathbf{z}^v$ be the fusion representation of $\{\mathbf{x}^v\}$ and the separation representation of each $\mathbf{x}^v$, respectively. Given $\mathbf{z}=\mathbf{z}^c+\{\mathbf{z}^v\}$, it is always possible to find a target $y$ for which $\mathbf{z}^c$ is not related to $y$ while $\mathbf{z}$ is, namely $\exists y.~I(y;\mathbf{z})>I(y;\mathbf{z}^c)=0$.

We now prove this theorem by construction.
\begin{proof}
Let $a$, $b$, and $c$ be any random variables, we first enumerate some basic properties of mutual information:

P1: $I(a;b)\ge0$, $I(a;b|c)\ge 0$ (\textit{positivity});

P2: $I(ab;c)\!=\!I(b;c)\!+\!I(a;c|b)$, $I(a;b;c)\!=\!I(b;c)\!-\!I(b;c|a)$  (\textit{chain rule}).

We make the following hypotheses:

H1: Since $\mathbf{z}^c$ is a representation of $\{\mathbf{x}^v\}$, $I(y;\mathbf{z}^c|\{\mathbf{x}^v\})=0$. 

H2: As $\mathbf{z}$ is a combination of $\mathbf{z}^c$ and $\{\mathbf{z}^v\}$, we have $I(\{\mathbf{x}^v\};\mathbf{z})>I(\{\mathbf{x}^v\};\mathbf{z}^c)$.

Then, we factorize $\{\mathbf{x}^v\}$ as a function $f(\cdot)$ of two independent random variables (according to \cite{DBLP:journals/jmlr/AchilleS18}) by picking $y$ such that:

C1: $I(y,\mathbf{z}^c)=0$, and C2: $\{\mathbf{x}^v\}=f(y,\mathbf{z}^c)$.

Since $\{\mathbf{x}^v\}$ is a function of $y$ and $\mathbf{z}^c$, C3: $I(\{\mathbf{x}^v\};\mathbf{z}|y\mathbf{z}^c)=0$.

We next formulate $I(y,\mathbf{z})$ as
\begin{equation*}
\setlength{\jot}{-2pt}
\begin{aligned}
    I(y;\mathbf{z}) &\overset{P2}{=} I(y;\mathbf{z}|\{\mathbf{x}^v\}) + I(\{\mathbf{x}^v\};y;\mathbf{z}) \\
     &\overset{P1}{\ge} I(\{\mathbf{x}^v\};y;\mathbf{z}) \\
     & \overset{P2}{=} I(\{\mathbf{x}^v\};\mathbf{z}) - I(\{\mathbf{x}^v\};\mathbf{z}|y) \\
     & \overset{P2}{=} I(\{\mathbf{x}^v\};\mathbf{z}) - I(\{\mathbf{x}^v\};\mathbf{z}|y\mathbf{z}^c) - I(\{\mathbf{x}^v\};\mathbf{z};\mathbf{z}^c|y) \\
     & \overset{C3}{=} I(\{\mathbf{x}^v\};\mathbf{z}) - I(\{\mathbf{x}^v\};\mathbf{z};\mathbf{z}^c|y) \\
     & \overset{P2}{=} I(\{\mathbf{x}^v\};\mathbf{z}) - I(\{\mathbf{x}^v\};\mathbf{z}^c|y) + I(\{\mathbf{x}^v\};\mathbf{z}^c|y\mathbf{z}) \\
     & \overset{P1}{\ge} I(\{\mathbf{x}^v\};\mathbf{z}) - I(\{\mathbf{x}^v\};\mathbf{z}^c|y) \\
     & \overset{P2}{=} I(\{\mathbf{x}^v\};\mathbf{z}) - I(\{\mathbf{x}^v\};\mathbf{z}^c) + I(\{\mathbf{x}^v\};y;\mathbf{z}^c) \\
     & \overset{P2}{=} I(\{\mathbf{x}^v\};\mathbf{z}) - I(\{\mathbf{x}^v\};\mathbf{z}^c) + I(y;\mathbf{z}) - I(y;\mathbf{z}^c|\{\mathbf{x}^v\}) \\
     & \overset{P1}{\ge} I(\{\mathbf{x}^v\};\mathbf{z}) - I(\{\mathbf{x}^v\};\mathbf{z}^c) - I(y;\mathbf{z}^c|\{\mathbf{x}^v\}) \\
     & \overset{H1}{=} I(\{\mathbf{x}^v\};\mathbf{z}) - I(\{\mathbf{x}^v\};\mathbf{z}^c) \\
     & \overset{H2}{>} 0.
     \end{aligned}
\end{equation*}

Since $I(y,\mathbf{z}^c)=0$ by construction and $I(y,\mathbf{z})>0$, the $y$ satisfies the results as shown in \textbf{Theorem 1}, i.e., $\exists y.~I(y;\mathbf{z})\!>\!I(y;\mathbf{z}^c)\!=\!0$.
\end{proof}

\end{document}